\documentclass{article}

\usepackage{PRIMEarxiv}

\usepackage[utf8]{inputenc} % allow utf-8 input
\usepackage[T1]{fontenc}    % use 8-bit T1 fonts
\usepackage{hyperref}       % hyperlinks
\usepackage{url}            % simple URL typesetting
\usepackage{booktabs}       % professional-quality tables
\usepackage{amsfonts}       % blackboard math symbols
\usepackage{nicefrac}       % compact symbols for 1/2, etc.
\usepackage{microtype}      % microtypography
\usepackage{lipsum}
\usepackage{fancyhdr}       % header
\usepackage{graphicx}       % graphics
\graphicspath{{media/}}     % organize your images and other figures under media/ folder
\usepackage{amsmath}
\usepackage{float}

\title{Fourier Neural Differential Equations for learning Quantum Field Theories
}

\author{
  Isaac Brant \\
  Department of Physics \\
  University of Cambridge \\
  \texttt{ib433@cantab.ac.uk} \\
   \And
  Alexander Norcliffe, Pietro Li\`o \\
  Department of Computer Science and Technology \\
  University of Cambridge \\
  \texttt{\{alin2, pl219\}@cam.ac.uk} \\
}

\begin{document}
\maketitle

\begin{abstract}
A Quantum Field Theory is defined by its interaction Hamiltonian, and linked to experimental data by the scattering matrix. The scattering matrix is calculated as a perturbative series, and represented succinctly as a first order differential equation in time. Neural Differential Equations (NDEs) learn the time derivative of a residual network’s hidden state, and have proven efficacy in learning differential equations with physical constraints. Hence using an NDE to learn particle scattering matrices presents a possible experiment-theory phenomenological connection. In this paper, NDE models are used to learn $\phi^4$ theory, Scalar-Yukawa theory and Scalar Quantum Electrodynamics. A new NDE architecture is also introduced, the Fourier Neural Differential Equation (FNDE), which combines NDE integration and Fourier network convolution. The FNDE model demonstrates better generalisability than the non-integrated equivalent FNO model. It is also shown that by training on scattering data, the interaction Hamiltonian of a theory can be extracted from network parameters.
\end{abstract}

% keywords can be removed

\section{Introduction}\label{sec:intro}
By multiplying the representations of the interaction vertices, propagators, and particle lines known as Feynman rules \cite{Feynman}, particle scattering amplitudes are derived from the interaction Hamiltonian. This is an example of a phenomenological connection between what is theoretically derived and what is empirically observed \cite{Heisenberg}. Neural Networks have been used to learn physical problems, including Hamiltonians \cite{Hamiltonian-NN}, higher-order behaviour \cite{SONODEs}, and Fourier representations \cite{FNN}, where physical constraints are applied to network architecture, improving convergence and explainability. Neural Differential Equations (NDEs) \cite{NODEs} take the continuous time limit of a Residual Neural Network (RNN) to learn differential equations. Integrating the learnt function through time outputs the network's hidden state to a continuous depth. NDEs have so far been applied to various quantum systems \cite{QNODEs-dissipation} \cite{QNODEs-QCD} \cite{QNODEs-qubits} \cite{lattice_qft_node}, but not yet to scattering processes in Quantum Field Theory (QFT).

In this paper, we look for how NDEs can be used to bridge the phenomenological connection between experiment and theory by training these models on particle scattering data to learn scalar quantum field theories. The objectives are twofold: \textbf{apply Neural Ordinary Differential Equations (NODE) to learn particle scattering}. \textbf{Introduce the Fourier Neural Differential Equation (FNDE)\footnote{Full source code for this paper can be found at \url{https://github.com/2357e2/fnde}}}, an NDE architecture designed to leverage second quantization.

\section{Theoretical Components}
\label{sec:theory}

\subsection{Neural ODEs}
A Neural Ordinary Differential Equation model \cite{NODEs} learns the velocity function of the network's hidden state and integrates the output with a numerical ODE solver. This is implemented by taking the limit of small step size of a residual network such that a hidden layer $\boldsymbol{z}_{t}$ with time discretization $t \in \{0, ..., T\}$ becomes $\boldsymbol{z}(t)$:
\begin{equation}\label{NODE}
    \frac{\mathrm{d}\boldsymbol{z}(t)}{\mathrm{d}t} = R(\boldsymbol{z}, \phi, t)
\end{equation}
where $R$ is the residual network taken to its continuous limit. This network architecture has several advantages over the discretized form \cite{NODEs}, including improved memory efficiency due to not backpropagating through the ODE solver, adaptation of temporal precision without retraining the model, and the model being continuous in time.

\subsection{The Scattering Matrix}\label{subsec:scattering}
In quantum mechanics, deviations from analytic potentials are encoded in the interaction Hamiltonian \cite{sakurai}.
\begin{equation}
    H = H_0 + H_{int}
\end{equation}
The scattering matrix element, $S_{fi}$, encodes the probability amplitude of transitioning from an initial configuration of particles/fields, $\vert\: i\: \rangle $, to a final configuration $\vert\: f\: \rangle $ via the interaction described by $H_{int}$:
\begin{equation}\label{Sfi}
    S_{fi} =  \langle \: f \: \vert  \: \psi(t) \: \rangle  = \langle \: f \: \vert \: \hat{S} \: \vert\: i\: \rangle
\end{equation}
where $\hat{S}$ is the Scattering operator. It is assumed that with sufficient separation, initial and final states are eigenstates of the free Hamiltonian. In the interaction picture, the interaction Hamiltonian becomes: ($\hat{H}_I = e^{i\hat{H}_0(t_1 - t_0)}\hat{H}_{int}e^{-i\hat{H}_0(t_1 - t_0)}$). Dyson's formula contains the time-ordering operator $\mathcal{T} $ \cite{Dyson}:
\begin{equation}\label{Dyson}
    S_{fi} = \langle \: f \: \vert \: \mathcal{T} \left\{ exp \left[ \frac{1}{i} \int_{-\infty}^{+\infty} \mathrm{d}t \hat{H_I}(t) \right] \right\} \: \vert \: i \: \rangle
\end{equation}
which is evaluated as a perturbative power series in time-ordered integrals of the interaction Hamiltonian. Each term of the series corresponds to the sum of Feynman diagrams with the number of vertices equal to the order of the term.

\subsection{The theory-experiment connection}
Experimental data from a particle collider experiment is contained within the differential cross section \cite{schwartz}:

\begin{equation}\label{cross}
    \mathrm{d}\sigma = \frac{1}{2E_{1}2E_{2} \vert \boldsymbol{v}_{1} - \boldsymbol{v}_{2}\vert} \vert \langle \: f \: \vert \: \mathcal{M} \: \vert\: i\: \rangle \vert ^2 \: \mathrm{d}\Pi    
\end{equation}

where $\mathrm{d}\Pi$ is an element of Lorentz-invariant phase space, and $\mathcal{M}$ is the Lorentz-invariant form of the Scattering matrix. $E_{1,2}$ and $\boldsymbol{v}_{1,2}$ are the kinematic variables of the incoming particles. The `theory' of the interaction is defined by the interaction Hamiltonian density, while the experimental corroboration is found in the cross section. Equation \eqref{cross} defines the cross section, while equation \eqref{Dyson} describes the involvement of the interaction Hamiltonian in a scattering process. The common element of both equations is the scattering matrix. This leads to the question: can a neural network learn the theory of particle interactions - from experimental data - via the scattering matrix?

To exploit the link between theory and experiment, the neural network must operate on scattering data but learn, indirectly, the interaction Hamiltonian. Hence the learning process should be split into two or more modules: one which learns the interaction Hamiltonian, and one which transforms the scattering data to an analogue of the Hamiltonian, and then back to the scattering data onto which the loss function and optimizer may act.

\section{NODE model}\label{sec:node}
Differentiating equation \eqref{Dyson} with respect to time leads to a straightforward relation:
\begin{equation}\label{eqn:dS}
    \frac{\mathrm{d}\hat{S}}{\mathrm{d}t} = \frac{1}{i}\hat{H}_{I}(t)\hat{S}(t)
\end{equation}
Apart from the operatorial nature, its form matches that of the NODE equation \eqref{NODE}. Hence by training a NODE network on scattering matrices, by design, the network has learnt the product of the interaction Hamiltonian and scattering operator for each time step. To perform numerical calculations with equation \eqref{eqn:dS}, a suitable representation must be applied. The most convenient method is to take matrix elements of the scattering operator between states $\vert\: i\: \rangle $ and $\vert\: f\: \rangle $.
\begin{equation}\label{dS_matrix}
\begin{split}
    \frac{dS_{fi}}{\mathrm{d}t} & = \frac{1}{i}\langle \: f \: \vert\hat{H}_{I}(t)\hat{S}(t) \vert\: i\: \rangle  = \frac{1}{i} \sum_{k} H_{I, fk}(t)S_{ki}(t)
\end{split}
\end{equation}
The interaction Hamiltonian is extracted from $R(\boldsymbol{z}, \phi, t)$ by post-multiplying by the inverse of $S$. 
\begin{equation}\label{H_I}
    H_{I, fi}(T) = \sum_{k}R_{fk}(\boldsymbol{z}, \phi, T)S^{-1}_{ki}(T)
\end{equation}

\section{Fourier NDE model}\label{sec:fnde}
Deriving $H_{I}$ from NODE form is a promising first attempt at deriving theory from experiment. But the quantity of interest to a field theorist is the interaction Hamiltonian density, $\mathcal{H}_I$, which has spatial dependence as well as temporal. Second-quantized operators in Quantum Field Theory are written as a Fourier expansion of creation and annihilation operators whose momentum basis defines that of the scattering matrix. Hence the required spatial dependence may be found within a Fourier transform. To do so we look for a new network architecture which combines an NDE with a Fourier Neural Operator Network \cite{Neural_Operator_PDEs}\cite{Fourier-PDE}. We look to make continuous the following form:
\begin{equation}\label{FNN_residual_form}
    \boldsymbol{z}_{t+1}(k) - \boldsymbol{z}_{t}(k) = \sigma \left\{W \boldsymbol{z}_{t}(k) + \mathcal{K}_{\phi} \boldsymbol{z}_{t}(k)\right\} - \boldsymbol{z}_{t}(k)
\end{equation}
where $W$ is a linear operator and $\mathcal{K}$ is the convolution operator. Taking the continuous limit in $t$, we get the Fourier Neural Differential Equation:
\begin{equation}\label{eqn:fnde}
\begin{split}
    \frac{\mathrm{d}\boldsymbol{z}(t, k)}{\mathrm{d}t} &= \sigma \left\{ W\boldsymbol{z}(t, k) + \mathcal{K}_{\phi} \boldsymbol{z}(t, k) \right\} - \boldsymbol{z}(t, k) \\
    &\equiv \sigma \left\{ W\boldsymbol{z}(t, k) + \mathcal{F}^{-1}\left[ \mathcal{F}(\kappa_{\phi})\cdot \mathcal{F}(\boldsymbol{z}) \right](t, k) \right\} - \boldsymbol{z}(t, k)
\end{split}
\end{equation}
The contraction with the transformed network kernel may be represented by a matrix multiplication, where the dimension of the kernel determines the filter's reciprocal space mode cutoff, interpreted as filtering spatial paths to those closest to the classical path.

\subsection{Modified FNDE form}

Now consider a modified form of equation \ref{eqn:fnde} which harmonises with the NODE form of equation \ref{eqn:dS}. Add the linear operator inside the Fourier transform rather than outside, which in turn can be combined with the subtracted factor of $\mathbf{z}(t, k)$ if the activation function is omitted.
\begin{equation}\label{FNDE_mod}
    \frac{\mathrm{d}\boldsymbol{z}(t, k)}{\mathrm{d}t} = \mathcal{F}^{-1}\left[ W\mathcal{F}(\boldsymbol{z})  + \mathcal{F}(\kappa_{\phi})\cdot \mathcal{F}(\boldsymbol{z})\right](t, k)
\end{equation}

This form may be applied to the scattering matrix to show explicit dependency on the interaction Hamiltonian density. The derivation can be found in appendix \ref{app:fnde_proof}, with the final result:

\begin{equation}\label{FNDE_interaction_density_result}
    \frac{\mathrm{d}S(t, p_f, p_i)}{\mathrm{d}t} = \mathcal{F}^{-1}\left\{e^{-i(\pi/2 + \boldsymbol{p}\cdot\boldsymbol{x})} \ \Bar{\mathcal{H}}_{I}(t, p_f, p_i, x_f, x_i) \mathcal{F} \left[ S(t, p_f, p_i)
    \right]\right\}\\
\end{equation}

With $\Bar{\mathcal{H}}_{I}$ the doubly-block circulant Toeplitz \cite{Toeplitz} form of the interaction Hamiltonian. For a scattering matrix of momentum discretization $n_p \times n_p$, this form of the interaction Hamiltonian will have dimension $n_p\times(\left \lfloor{n_p/2}\right \rfloor + 1)$. It can be identified from the network kernel and linear operator:
\begin{equation}
    \left[W + \mathcal{F}(\kappa_{\phi})\right]_{fi} = e^{-i (\pi/2 +\boldsymbol{p}\cdot\boldsymbol{x})}\ \Bar{\mathcal{H}}_{I, fi}(t, x_f, x_i)
\end{equation}

\section{Experiments and Results}
\subsection{Experimental setup}
Obtaining the interaction Hamiltonian is the primary purpose of these NDE models, rather than quicker convergence to the scattering matrix. Hence other models - such as MLPs, Residual networks - are not tested. NODE (equation \ref{eqn:dS}), FNDE (equation \ref{eqn:fnde}), modified FNDE (equation \ref{FNDE_interaction_density_result}) and FNO \cite{Fourier-PDE} models are compared against their ability to learn Scalar Quantum Electrodynamics (QED) \cite{schwartz}, Scalar Yukawa \cite{scalar-Yukawa}, and $\phi^4$ theories \cite{peskin} (see appendix \ref{app:scalar}). To roughly equalise training times, the NODE model contained 3 layers, while the Fourier models contained a single Fourier layer. Unless otherwise specified, all tests used the same 400 epoch training loop with an Adam optimizer \cite{adam} of initial learning rate 0.02 which was halved after 100 and 250 epochs. The batch size was 16 to vary four values of the coupling constant with four values of the particle mass. Integration was handled by torchdiffeq \cite{torchdiffeq} using a fourth-order Runge-Kutta \cite{runge_kutta} numerical integrator applied over 10 time steps. All tests were run five times and the mean taken.

\subsection{Convergence tests}
\begin{figure}[ht]
    \centering
    \includegraphics[width=54mm, height=42mm]{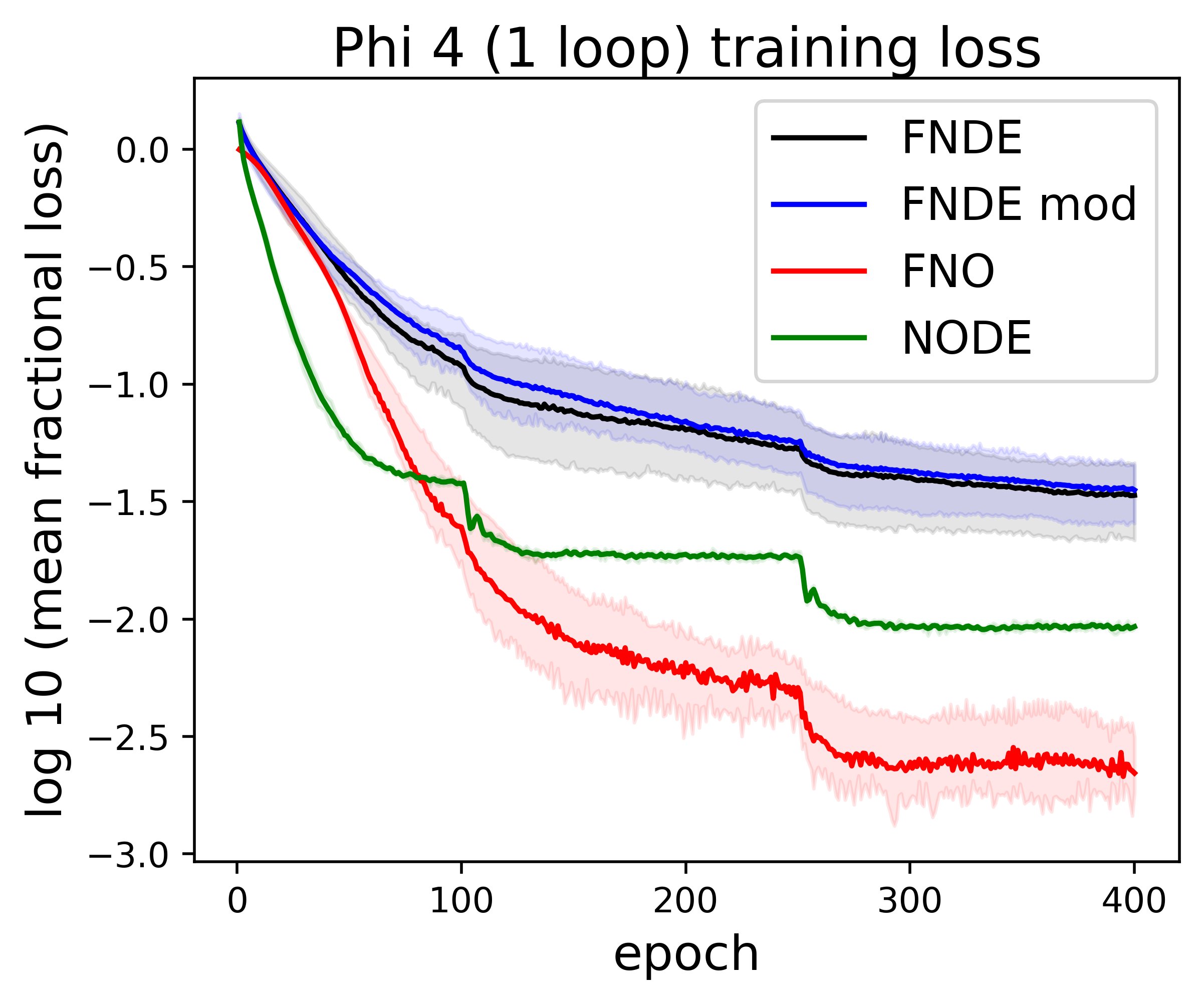}
    \includegraphics[width=54mm, height=42mm]{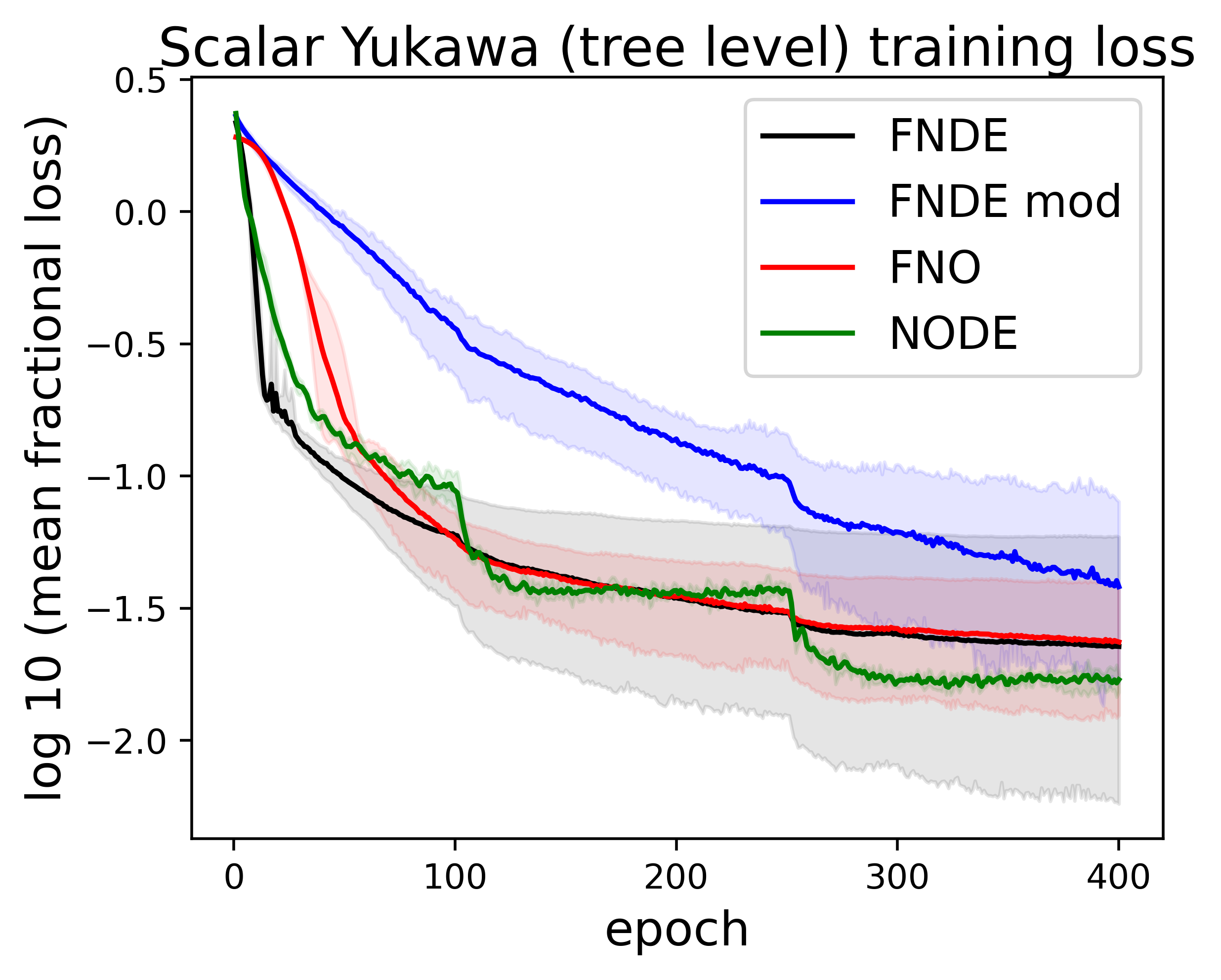}
    \includegraphics[width=54mm, height=42mm]{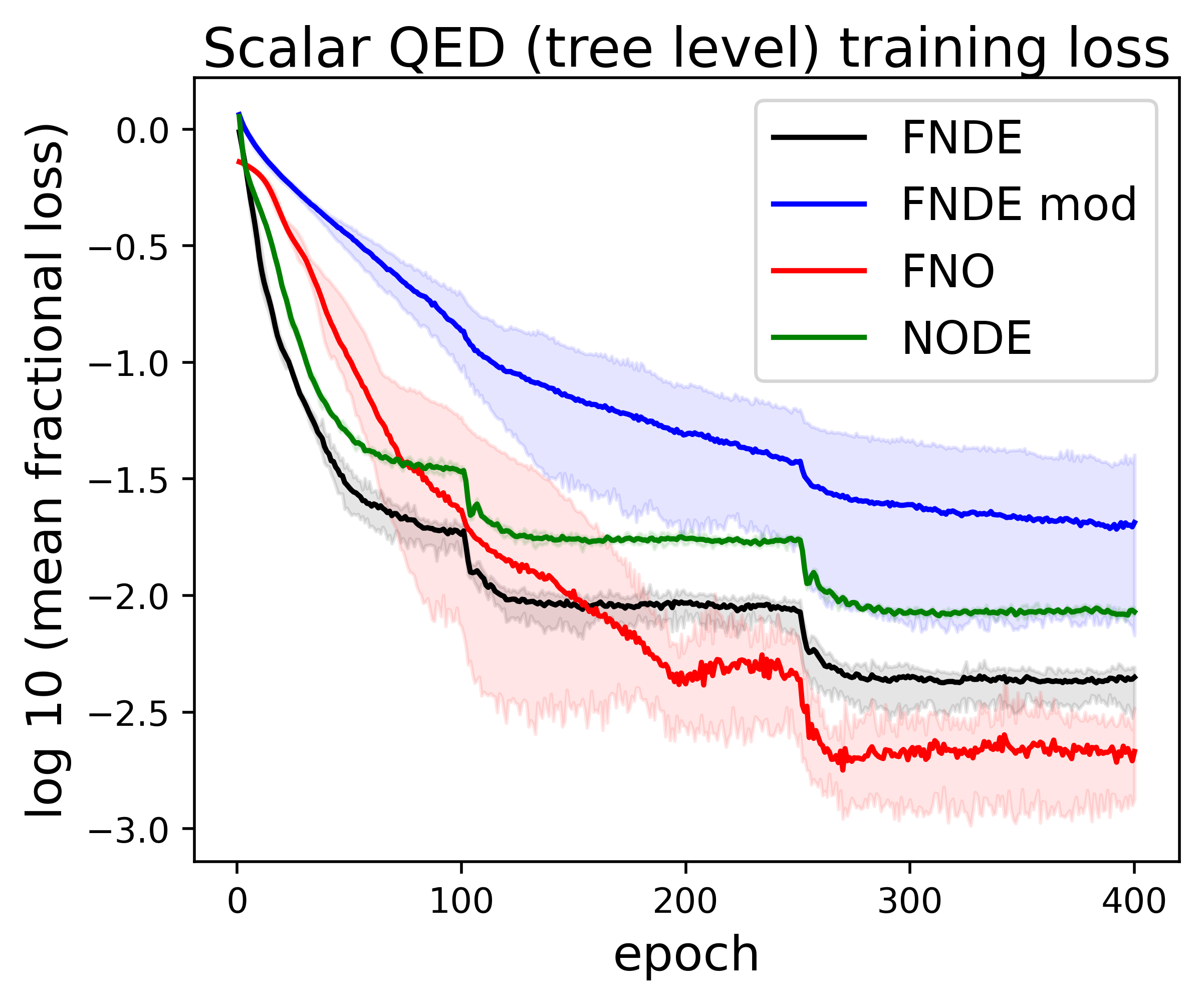}
    \caption{Training losses for the FNDE, modified FNDE, FNO and NODE models. There is some inconsistency between model performances, but overall the modified FNDE performs worst on learning training data, while the FNO model performs consistently well. Note the drop in loss after 100 and 250 epochs corresponding to the decrease in learning rate.}
    \label{fig:training_loss}
\end{figure}

\begin{figure}[H]
    \centering
    \includegraphics[width=54mm, height=42mm]{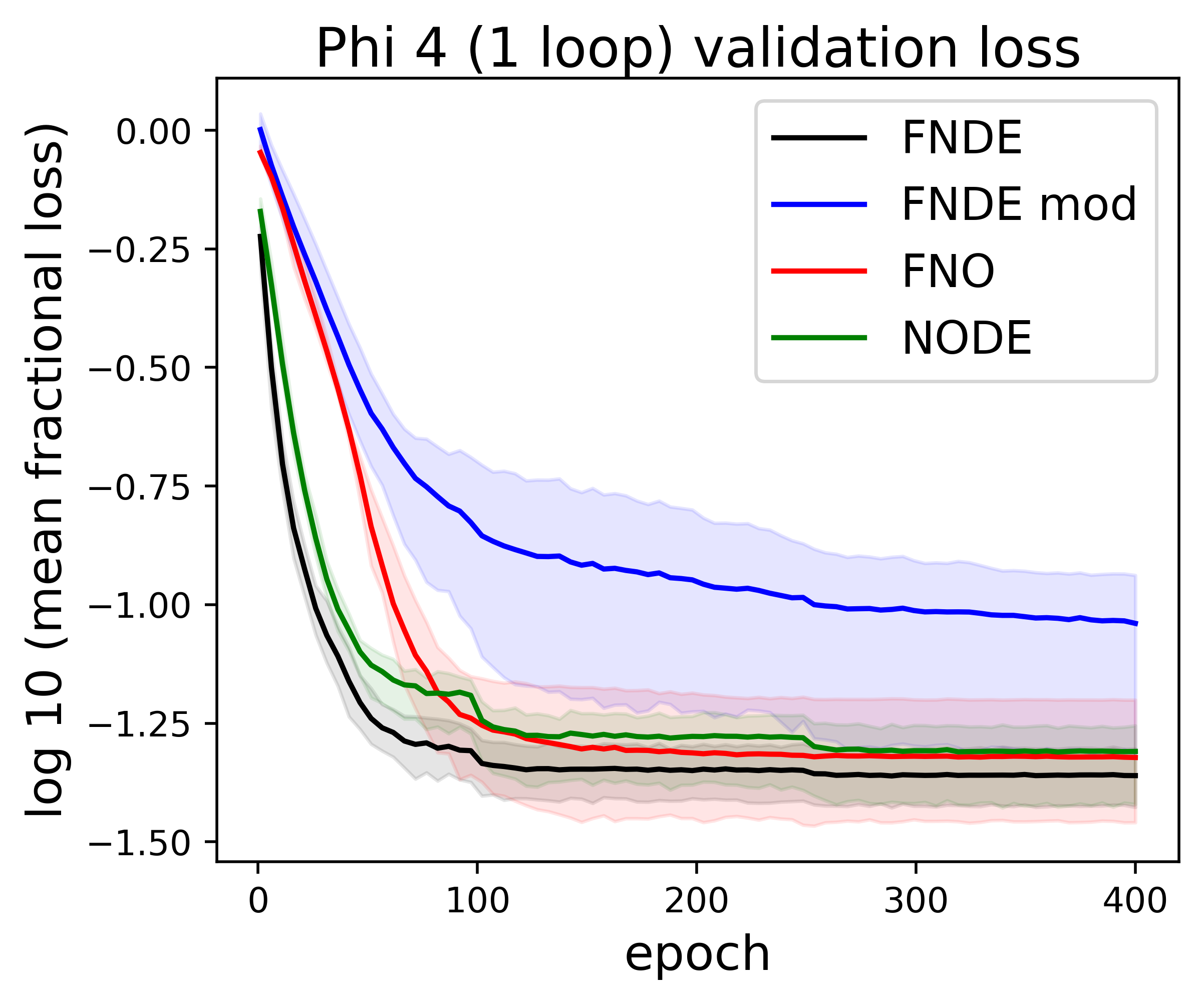}
    \includegraphics[width=54mm, height=42mm]{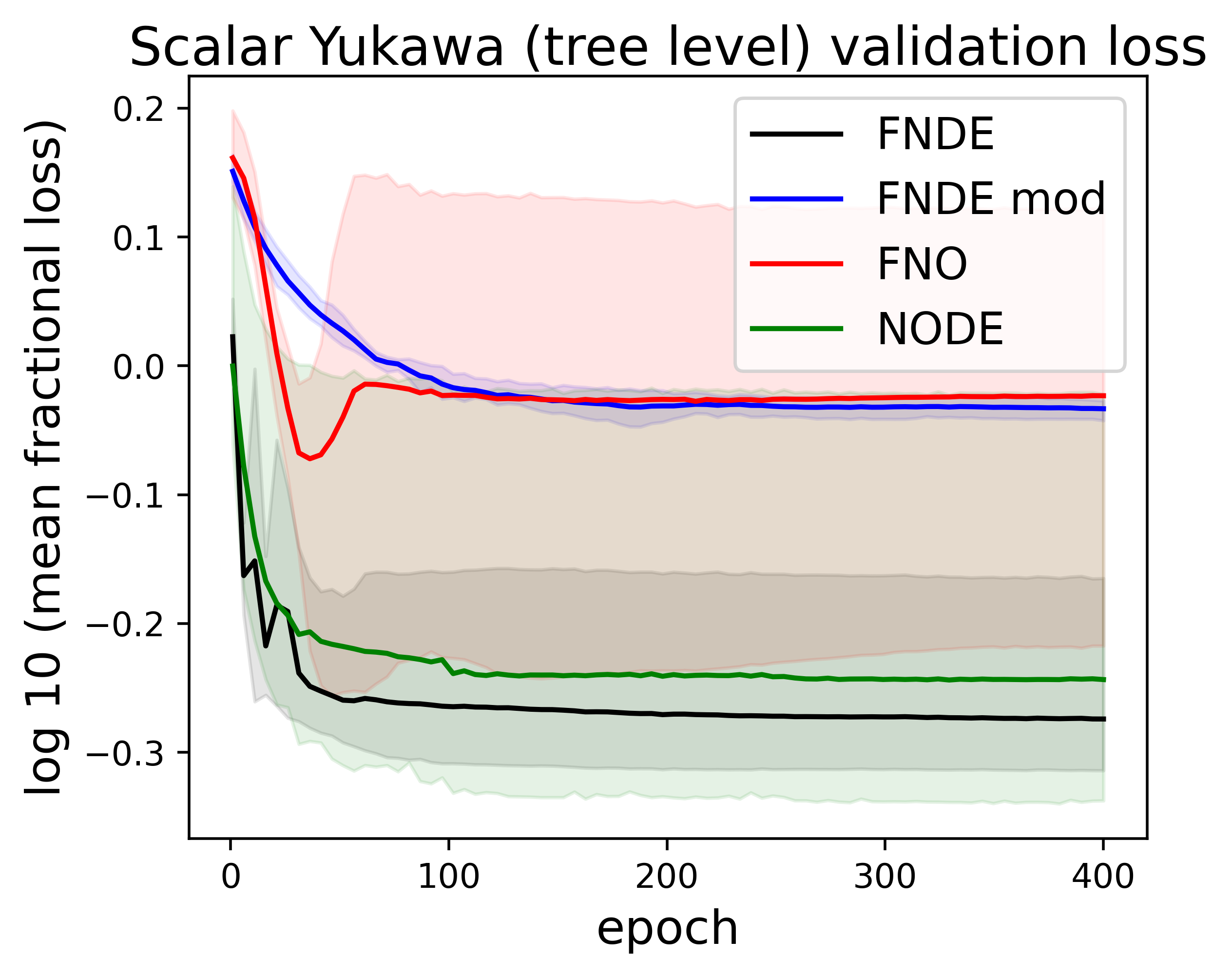}
    \includegraphics[width=54mm, height=42mm]{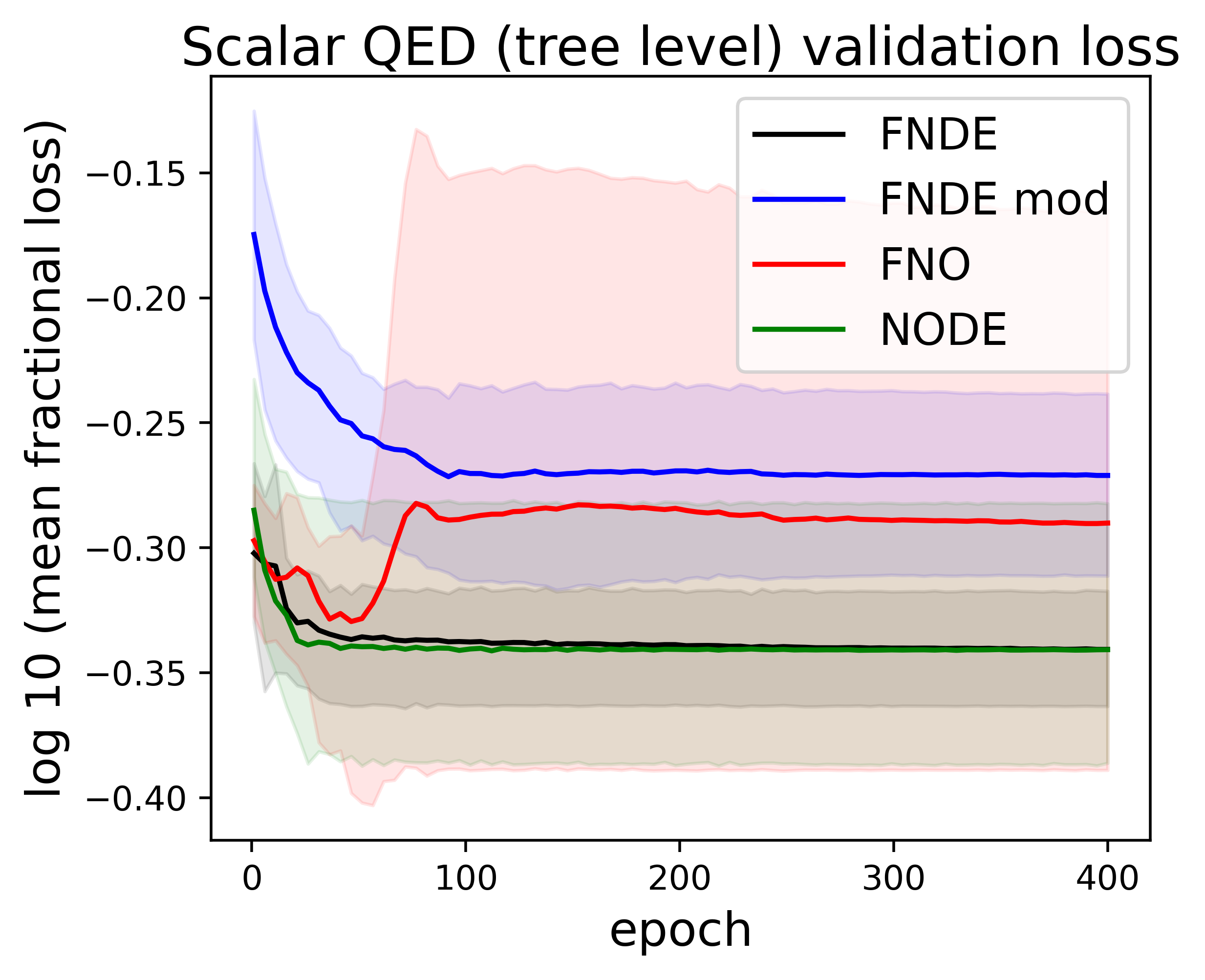}
    \caption{Validation losses during training for FNDE, modified FNDE, FNO and NODE models. This tests the ability to generalise to interpolated momentum discretizations. The FNDE maintains a slight overall lead.}
    \label{fig:validation_data}
\end{figure}

The first test, in figures \ref{fig:training_loss} and \ref{fig:validation_data}, assessed each model's ability to converge to the analytic value of the scattering matrix.

Figure \ref{fig:validation_data} suggests certain theories lend themselves to generalisation more than others. For NDE models, the $\phi^4$ theory's validation loss doesn't differ too far from its training loss, whereas the other two theories have a much lower validation loss than in the training set. This failure to generalise is particularly apparent in the FNO model, suggesting overfitting, whereas the NDE's integration seems to improve generalisability.

\subsection{Higher-order diagrams}
\begin{figure}[H]
    \centering
    \includegraphics[width=54mm, height=42mm]{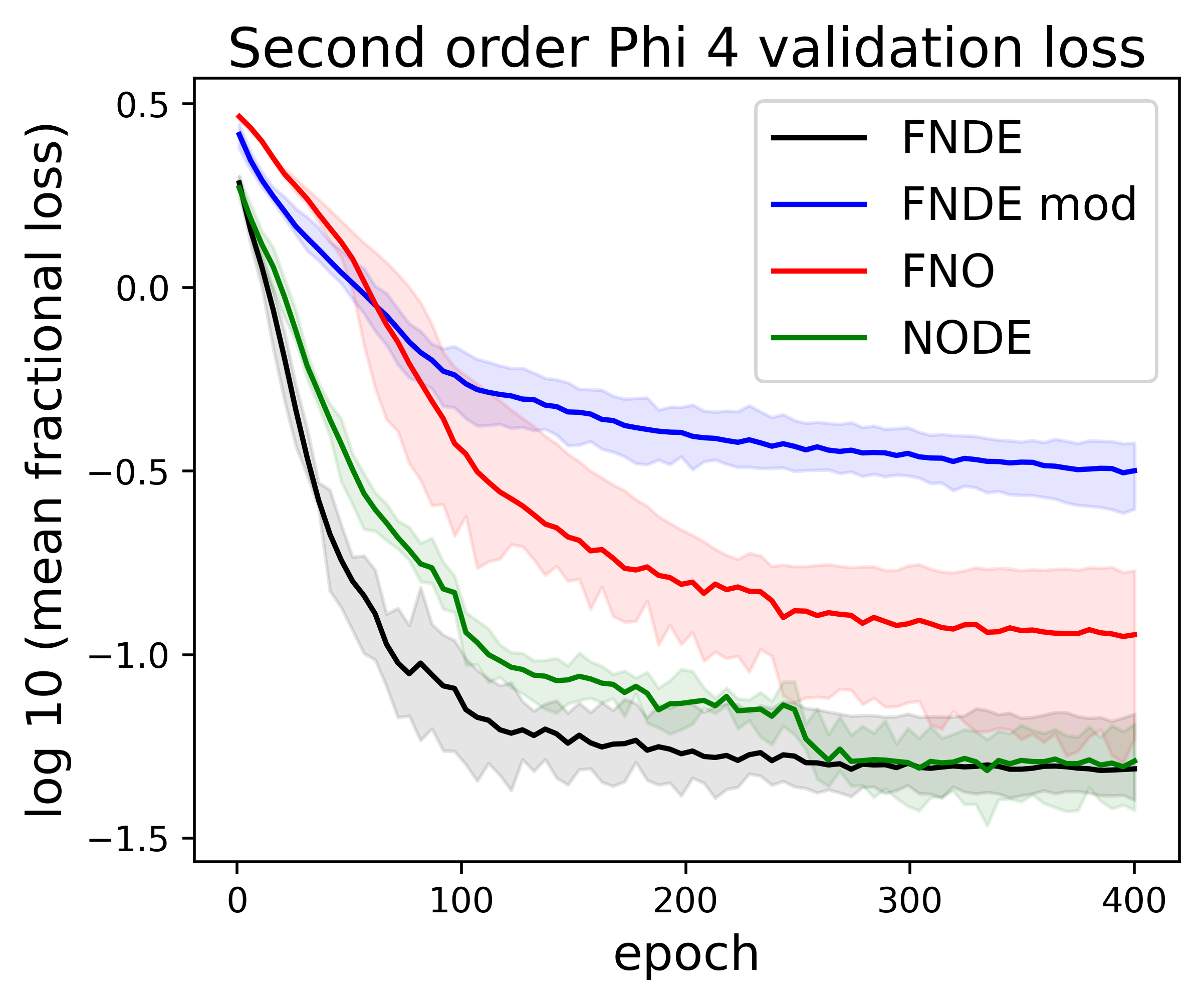}
    \includegraphics[width=54mm, height=42mm]{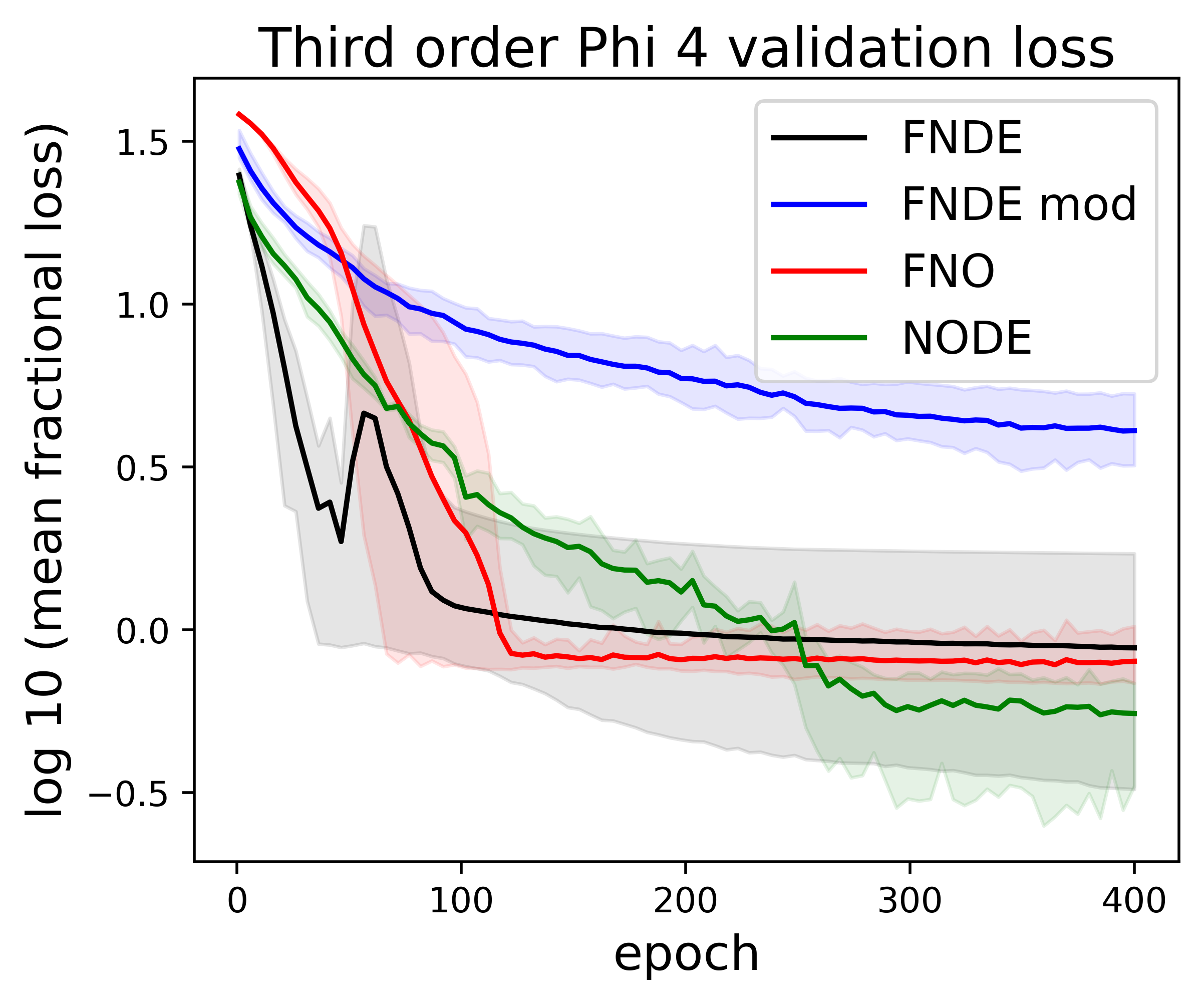}
    \caption{Validation losses for second and third order $\phi^4$ theory terms. Higher order terms become more significant with increasing interaction energy.}
    \label{fig:loop_losses}
\end{figure}

The utility of a model learning perturbation theory is limited by its ability to discern higher order terms/diagrams in the expansion. In figure \ref{fig:loop_losses}, the models are tested in their ability to learn the terms in the perturbation which are of larger powers in the coupling constant, corresponding to Feynman diagrams with greater vertex number.

Higher-order terms are less linear, and often require greater model complexity to analogise. Again, the modified FNDE model suffers, but the other three maintain similar performance, with the node model having a slight edge on the third-order terms.

Additional results can be found in appendix \ref{app:experiments}.

\section{Discussion and Conclusions}
NDEs have been demonstrated to learn Quantum Field Theories from scattering data, with the interaction Hamiltonian (density) extricable from the trained network. The Fourier Neural Differential Equation was derived and motivated, with its performance compared to other models. All models show acceptable convergence to lower-order diagrams, with the FNDE model providing a slight overall advantage in performance. Convergence to higher-order diagrams could be improved with increased depth and integration steps. While the non-integrated FNO model demonstrated better convergence to training data, the NDE models demonstrated superior performance on the validation data, suggesting better generalisability. The primary result remains the theoretical demonstration of how quantities of interest can be indirectly learnt from network parameters; network architectures can be designed to leverage physical correspondences.

\subsection{Further Development}
\begin{itemize}
    \item \textbf{Refining FNDE structure}  Further tuning of the Fourier layer structure and adding multiple integration layers in sequence could improve convergence and range of applications to problems involving time-dependent Fourier decomposition. Additional testing could demonstrate the circumstances under which it provides a computational advantage. The modified form the FNDE suffered in performance compared to the unmodified FNDE. Deriving a form of the interaction Hamiltonian density from the unmodified FNDE could combine theoretical and performance advantages.
    \item \textbf{Extracting symbolic form of the interaction Hamiltonian}  Given it is possible to find a numerical form, the natural next step is to attempt to find the symbolic form, such as by using symbolic regression. If effective, this could be a useful tool for particle physicists to extract theory from experimental data.
\end{itemize}

%Bibliography
\bibliographystyle{unsrt}  
\bibliography{fnde}  

\section*{Appendices}
\appendix
\section{Scattering matrix form of FNDE derivation}\label{app:fnde_proof}
A scattering matrix element is function of both initial and final momenta, so we can write
\begin{equation}\label{dS_mod}
\begin{split}
    \frac{\mathrm{d}S(t, p_f, p_i)}{\mathrm{d}t} &= \mathcal{F}^{-1}\left[ W\mathcal{F}(S)  + \mathcal{F}(\kappa_{\phi})\cdot \mathcal{F}(S)\right](t, p_f, p_i) \\
    &= \mathcal{F}^{-1}\left\{\left[ W  + \mathcal{F}(\kappa_{\phi})\right] \cdot \mathcal{F}(S)\right\}(t, p_f, p_i) \\
    &= \frac{1}{i} \sum_{k} H_{I}(t, p_f, p_k)S(t, p_k, p_i)
\end{split}
\end{equation}
A discrete 2d convolution may be written as a matrix multiplication where the kernel of convolution becomes the doubly-block circulant (Toeplitz) form of itself \cite{Toeplitz}. To perform this operation backwards, we define $\Bar{H}_I$ such that $H_I$ is the doubly-block circulant form of $\Bar{H}_I$. Hence we may rewrite our FNDE in terms of this two dimensional discrete convolution.
\begin{equation}\label{Toeplitz}
\begin{split}
    \frac{\mathrm{d}S(t, p_f, p_i)}{\mathrm{d}t} &= \frac{1}{i} \left[ \Bar{H}_{I}(t, p_f, p_i) * S(t, p_f, p_i) \right] \\
    &= \mathcal{F}^{-1}\left\{ \mathcal{F}\left[ \frac{1}{i}\Bar{H}_{I}(t, p_f, p_i) * S(t, p_f, p_i)
    \right]\right\} \\
    &= \mathcal{F}^{-1}\left\{ \mathcal{F}\left[ \frac{1}{i}\Bar{H}_{I}(t, p_f, p_i)\right] \mathcal{F} \left[ S(t, p_f, p_i)
    \right]\right\}
\end{split}
\end{equation}
By comparing \eqref{dS_mod} and \eqref{Toeplitz}, we can identify the Fourier transform of the interaction Hamiltonian's Toeplitz form matrix element with the linear operator and transformed kernel. Hence these network parameter values can be used to reconstruct a numerical form of the interaction Hamiltonian. Compared to the NODE model described in section \ref{sec:node}, we have a similar expression for the interaction Hamiltonian in terms of learnt network parameters. But apart from some conjecturing about Fourier filtering, there is no \emph{a priori} advantage over the NODE model. But if we write the interaction Hamiltonian as a spatial integral over the interaction Hamiltonian density, the Fourier transform and spatial integral cancel except for an exponential phase term.

\begin{equation}\label{FNDE_interaction_density}
\begin{split}
    \frac{\mathrm{d}S(t, p_f, p_i)}{\mathrm{d}t} &=  \mathcal{F}^{-1}\left\{ \mathcal{F}\left[ \frac{1}{i}\Bar{H}_{I}(t, p_f, p_i)\right] \mathcal{F} \left[ S(t, p_f, p_i)
    \right]\right\}\\
    &= \mathcal{F}^{-1}\left\{ \mathcal{F}\left[ \int d^2x \ \frac{1}{i}\Bar{\mathcal{H}}_{I}(t, p_f, p_i, x_f, x_i)\right] \mathcal{F} \left[ S(t, p_f, p_i)
    \right]\right\}\\
    &= \mathcal{F}^{-1}\left\{ \int \mathrm{d}^2 p \ e^{ - i \boldsymbol{p}\cdot\boldsymbol{x}} \left[ \int \mathrm{d}^2x \ \frac{1}{i}\Bar{\mathcal{H}}_{I}(t, p_f, p_i, x_f, x_i)\right] \mathcal{F} \left[ S(t, p_f, p_i)
    \right]\right\}\\
    &= \mathcal{F}^{-1}\left\{ \int \mathrm{d}^2 p \ e^{ - i \boldsymbol{p}\cdot\boldsymbol{x}} \left[ \int \mathrm{d}^2x \ e^{i \boldsymbol{p}\cdot\boldsymbol{x}} e^{-i \boldsymbol{p}\cdot\boldsymbol{x}} \ \frac{1}{i}\Bar{\mathcal{H}}_{I}(t, p_f, p_i, x_f, x_i)\right] \mathcal{F} \left[ S(t, p_f, p_i)
    \right]\right\}\\
    &= \mathcal{F}^{-1}\left\{e^{-i(\pi/2 + \boldsymbol{p}\cdot\boldsymbol{x})} \ \Bar{\mathcal{H}}_{I}(t, p_f, p_i, x_f, x_i) \mathcal{F} \left[ S(t, p_f, p_i)
    \right]\right\}\\
\end{split}
\end{equation}
Hence we can make another comparison between the network kernel and the Hamiltonian, but in this instance it is the interaction Hamiltonian density - an advantage over the NODE model.
\begin{equation}
    \left[W + \mathcal{F}(\kappa_{\phi})\right]_{fi} = e^{-i (\pi/2 +\boldsymbol{p}\cdot\boldsymbol{x})}\ \Bar{\mathcal{H}}_{I, fi}(t, x_f, x_i)
\end{equation}
As the operator and FFT of the kernel are defined within reciprocal space, they are spatially dependent - as required to get the Hamiltonian density. For a scattering matrix of momentum discretization $n_p \times n_p$, the resulting Toeplitz interaction Hamiltonian density will be of dimension $n_p\times(\left \lfloor{n_p/2}\right \rfloor + 1)$.

\section{Scalar Field Theories}\label{app:scalar}
\begin{figure}[H]
    \centering
    \includegraphics[width=120mm]{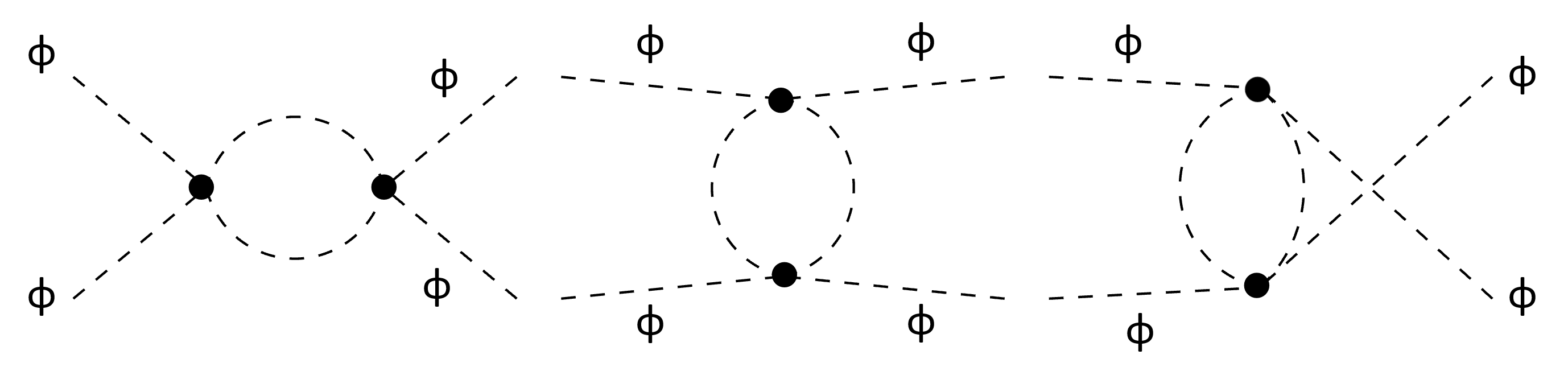}
    \caption{The 1-loop connected diagrams of $\phi^4$ theory. Generated with the Mathematica package FeynArts \cite{Feynarts}.}
    \label{fig:phi4}
\end{figure}

Known scattering matrices, calculated analytically, are used as model training and validation data. Besides momentum, the free parameters are particle mass and the coupling constant. These are varied to ensure the model is capable of handling a range of interactions. The chosen theories are evaluated at differing orders of perturbative evaluation. The selected test theories are:
\begin{itemize}
    \item $\phi^4$ theory, $ \mathcal{H}_{int} = \lambda \phi^{4}$ (figure \ref{fig:phi4})
    \item Scalar Yukawa theory, $\mathcal{H}_{int} = \lambda \phi \psi^\dagger \psi$
    \item Scalar QED, $\mathcal{H}_{int} = i\lambda \phi^\dagger(\partial_\mu \phi) A^\mu -i\lambda(\partial_\mu \phi^\dagger) \phi A^\mu -\lambda^2\phi^\dagger \phi A_\mu A^\mu -\lambda^2 \phi^\dagger\phi (A^0)^2$
\end{itemize}
where $\phi$ is a real scalar field, $\psi$ is a complex scalar field, $A^{\mu}$ is an Abelian gauge field, and $\lambda$ is the coupling constant. Scalar quantum field theories are preferable for development purposes over spinor theories to avoid the practical difficulties of four-component objects and single occupancy requirements in the implementation. Besides the theory itself, physical parameters may be varied, including the particle mass, the coupling constant, the momenta range, and the included Feynman diagrams.

The scattering matrix for these processes was synthetically generated\footnote{The explicit form of scattering matrix elements comes from \cite{peskin}, \cite{schwartz}, \cite{scalar-Yukawa}} to allow the momentum discretization and other parameters to be varied when testing the neural network models. Further studies could supplement synthetic data with real experimental data  processed to abide by a usable format.

\section{Additional experiments}\label{app:experiments}
\subsection{Extrapolating to higher energy scales}
Another potential use of this model is to make predictions about scattering experiment outcomes above the energy scales of our current collider experiments. To test this capability, a model was offered a momentum range extending beyond that on which it was trained, and the predicted output was compared to the analytic result.

\begin{figure}[H]
    \centering
    \includegraphics[width=120mm]{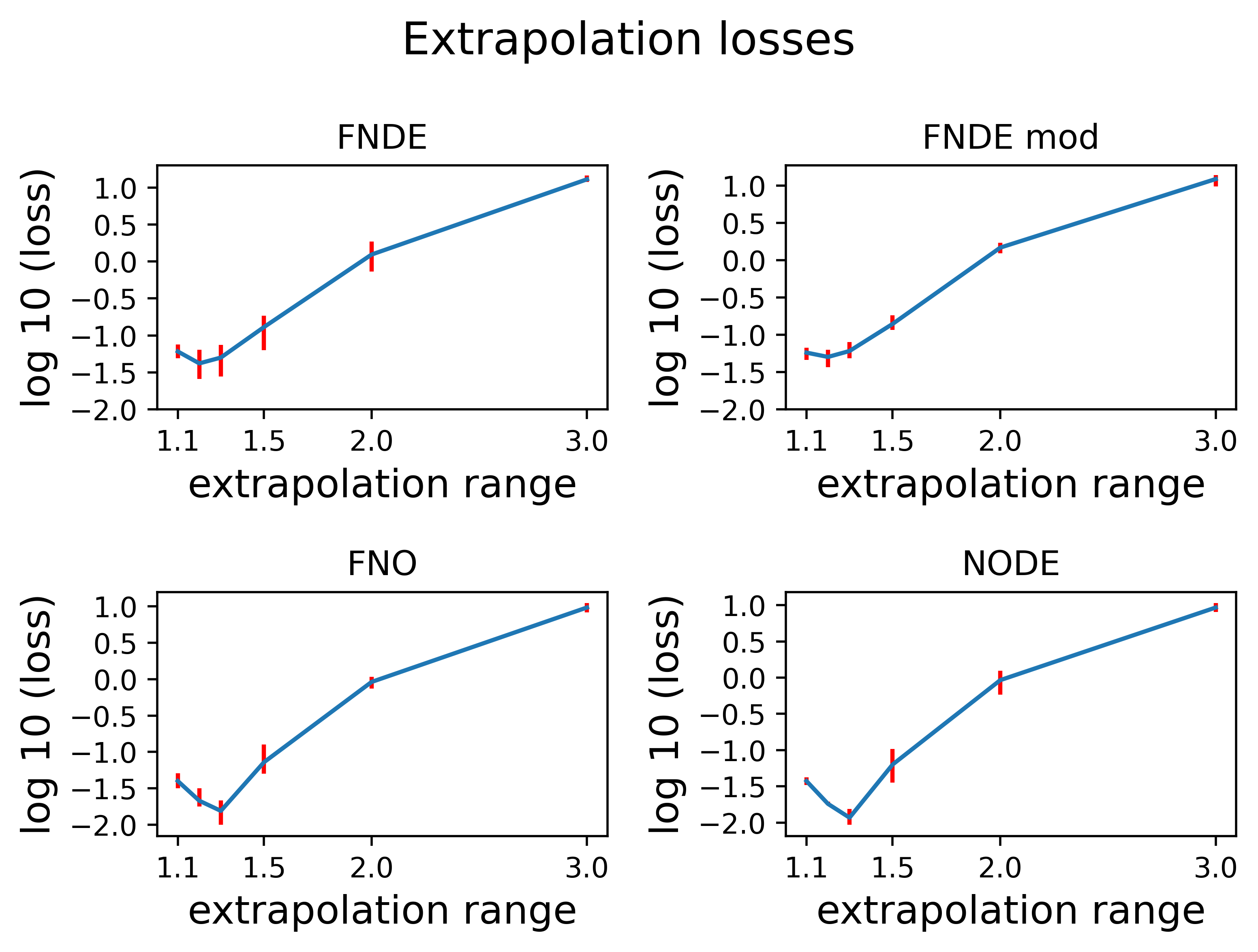}
    \caption{Mean fractional loss when predicting a scattering matrix outside the momentum range of a pre-trained $\phi^4$ model. The extrapolation range is ratio of the largest momentum in the test set to the training set. All models demonstrate acceptable losses up to 1.5 times the training momentum range. Error bars depict the range of losses for 5 trained models.}
    \label{fig:extrap}
\end{figure}

The extrapolation losses in figure \ref{fig:extrap} demonstrate all models are suitable for making predictions outside their immediate momentum range, but as expected, diverge from analytic results when the range extends too high, in this case from 1.5 times their initial training range. All models have the lowest loss at 1.2 or 1.3 times the initial momentum range, which is unexpected, and possibly caused by the input momentum matrix at 1.2/1.3 times lining up better with the input of the training data.

\subsection{Comparison of momentum discretizations and dimensional reduction}
\begin{figure}[H]
    \centering
    \includegraphics[width=54mm, height=42mm]{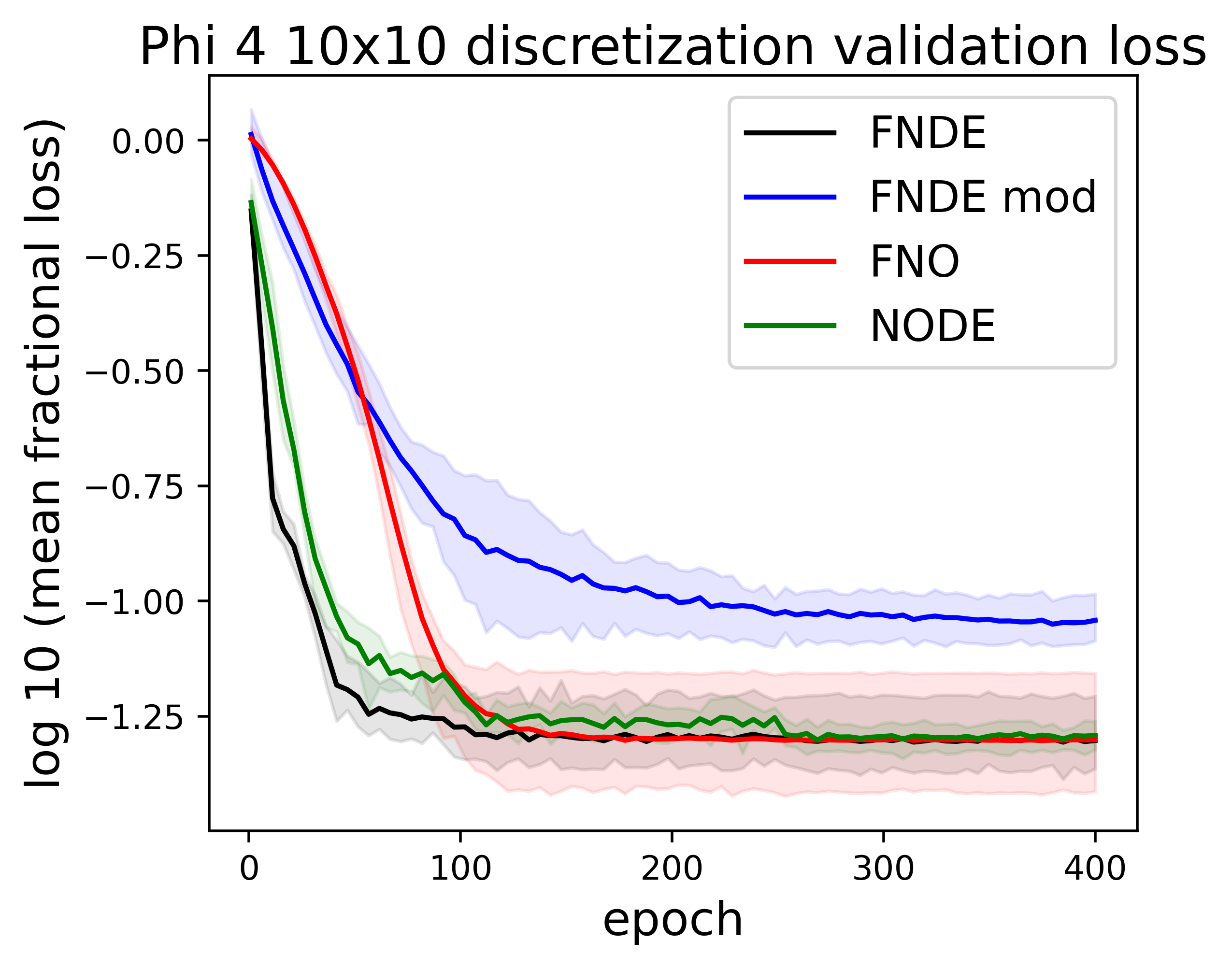}
    \includegraphics[width=54mm, height=42mm]{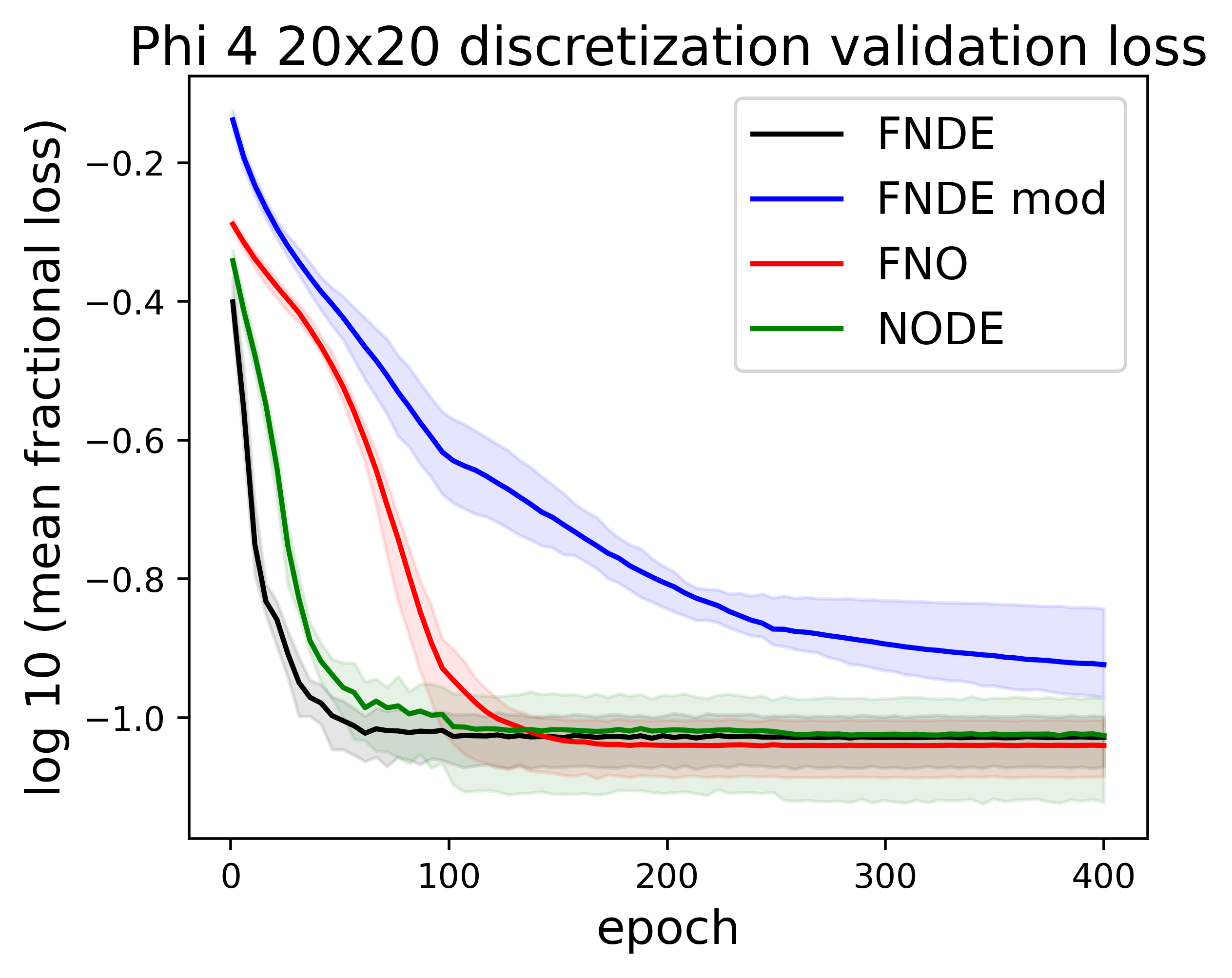}
    \includegraphics[width=54mm, height=42mm]{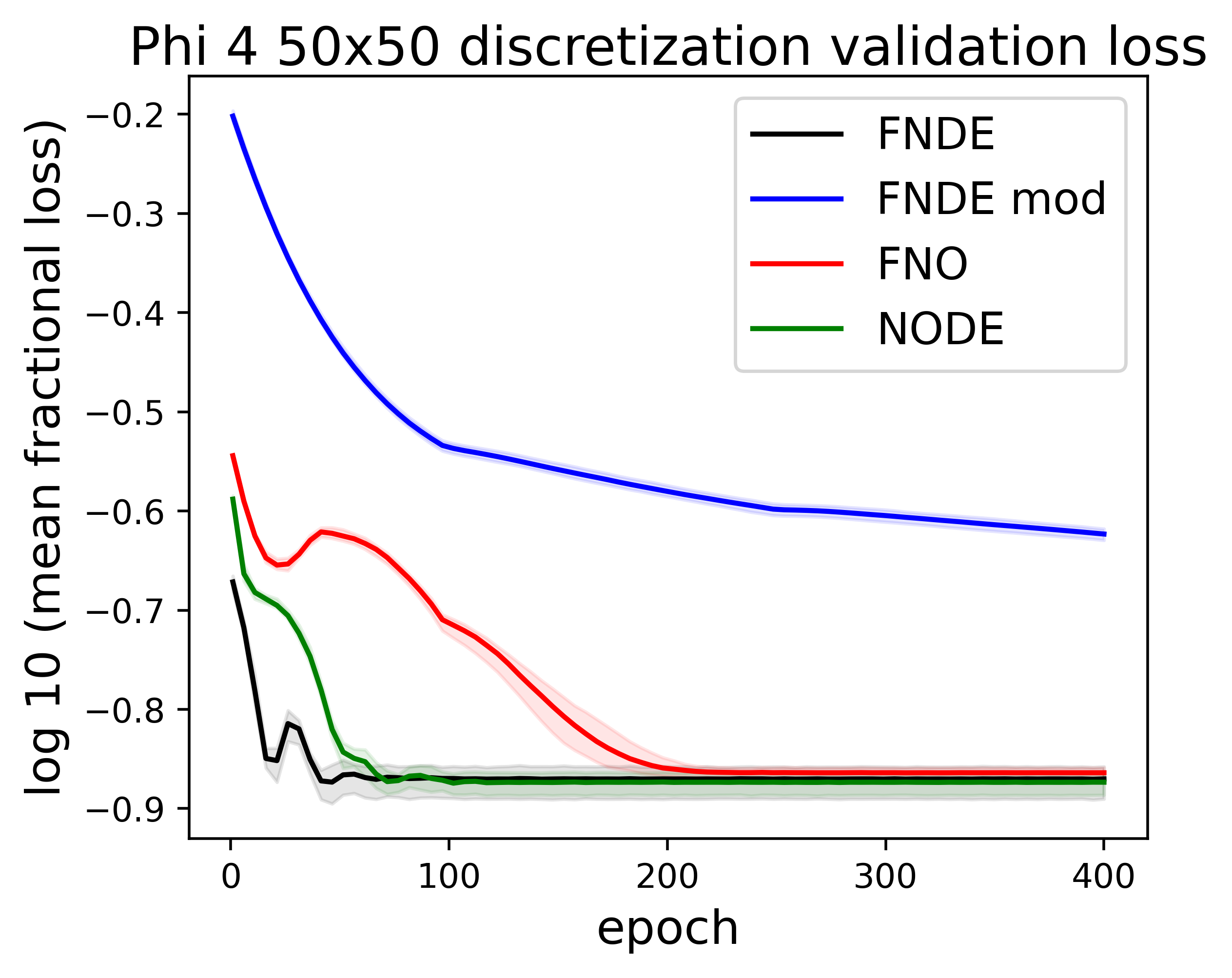}
    \caption{Validation losses for different momentum discretizations. Trained on 1 loop phi 4 theory with the same momentum range, but different precisions. Performance does drop with finer discretizations, but FNDE, NODE and FNO models converge to the same loss, albeit at different rates. }
    \label{fig:momentum_dis}
\end{figure}

However well a model may learn the scattering matrix for a specific momentum discretization, its use is very limited if, given an experimental data, it is unable to adapt to a different discretization. To demonstrate the functionality for different momentum discretizations, the validation losses were plotted in figure \ref{fig:momentum_dis} for momentum discretizations between a 100 element, a 400 element, and a 2500 element scattering matrix for single loop phi 4 theory.

Although performance drops slightly from the 10x10 to 50x50 discretization - as would be expected as the computational complexity increases quadratically - the models are still capable of converging to a reasonable extent without additional layers being introduced. Due to the large input size, all models introduce a `bottleneck' through dimensional reduction from the input layer size down to a hidden layer size of 100 in the case of the NODE model, and a Fourier-filtered size of 32 in the case of the Fourier models. Thus these models produce a well-performing dimensionally-reduced representation of the scattering matrix, implying their capacity to intuit the analytic scattering functions. The FNDE model has a slight edge in converging the quickest, suggesting a performance benefit when learning larger discretizations with shorter training times.

\end{document}